\documentclass[runningheads]{llncs}
\usepackage{graphicx}
\usepackage{amsmath,amssymb} % define this before the line numbering.
\usepackage{color}
\usepackage{makecell}
\usepackage[linesnumbered,ruled,vlined]{algorithm2e}%[ruled,vlined]{  
\usepackage{algpseudocode}  
\usepackage{amsmath}  
\usepackage{epstopdf} 
\DeclareUnicodeCharacter{2212}{\pm}
\begin{document}
\pagestyle{headings}
\mainmatter

%===========================================================
\title{SimpleTrack: Rethinking and Improving the JDE Approach for Multi-Object Tracking} % Replace with your title
\titlerunning{SimpleTrack}
\renewcommand{\thefootnote}{0}
\author{Jiaxin Li\inst{1} \and
Yan Ding\inst{1*} \and
Hua-Liang Wei\inst{2}}
\authorrunning{Jiaxin Li et al.}
\institute{Key Laboratory of Dynamics and Control of Flight Vehicle, Ministry of Education, School of Aerospace Engineering, Beijing Institute of Technology, Beijing 100081, China
\\
\and
Department of Automatic Control and Systems Engineering, University of Sheffield, Sheffield S1 3JD, UK
\footnotetext[0]
{
*Yan Ding is the corresponding author.
}
}
\maketitle
%===========================================================
\begin{abstract}
Joint detection and embedding (JDE) based methods usually estimate bounding boxes and embedding features of objects with a single network in Multi-Object Tracking (MOT). In the tracking stage, JDE-based methods fuse the target motion information and appearance information by applying the same rule, which could fail when the target is briefly lost or blocked. To overcome this problem, we propose a new association matrix, the Embedding and Giou matrix, which combines embedding cosine distance and Giou distance of objects. To further improve the performance of data association, we develop a simple, effective tracker named SimpleTrack, which designs a bottom-up fusion method for Re-identity and proposes a new tracking strategy based on our EG matrix. The experimental results indicate that SimpleTrack has powerful data association capability, e.g., 61.6 HOTA and 76.3 IDF1 on MOT17. In addition, we apply the EG matrix to 5 different state-of-the-art JDE-based methods and achieve significant improvements in IDF1, HOTA and IDsw metrics, and increase the tracking speed of these methods by about 20\%. 

\end{abstract}

%===========================================================
\section{Introduction}
Multi-object tracking (MOT), aiming to estimate the locations and identity of multiple targets in a video sequence, is a fundamentally challenging task in computer vision\cite{vandenhende2021multi}. Recently, the Intersection over Union (IoU) and Hungarian method are commonly used in the tracking phase among many tracking-by-detection paradigm\cite{bewley2016simple,bochinski2017high,liu2020gsm,specker2021occlusion,tang2017multiple,wojke2017simple,xiao2018simple,xu2019spatial,zhang2021bytetrack}. However, when the target is occluded or lost for a period of time, it is difficult to retrieve the correct identity only using the IoU distance. As a result, the identity switching of targets occurs from time to time. To alleviate this problem, many methods start to introduce the Re-identity feature of targets. Among them, the JDE-based methods\cite{liang2020rethinking,liang2021one,lu2020retinatrack,pang2021quasi,wang2020towards,zhang2021voxeltrack,zhang2021fairmot} have become popular due to their simplicity and efficiency. 

In the part of the data association, the accuracy of similarity measurement determines the tracking performance. Most detection-based methods use the IoU distance as the similarity matrix in the cascade matching strategy, while JDE-based methods fuse the motion information and appearance information as the similarity matrix for the linear assignment in the first matching and use the IoU distance in the next matching. However, none of these existing methods is the best expression of the similarity matrix according to our experiments. 
\begin{figure}
\centering
\includegraphics[height=3.2cm]{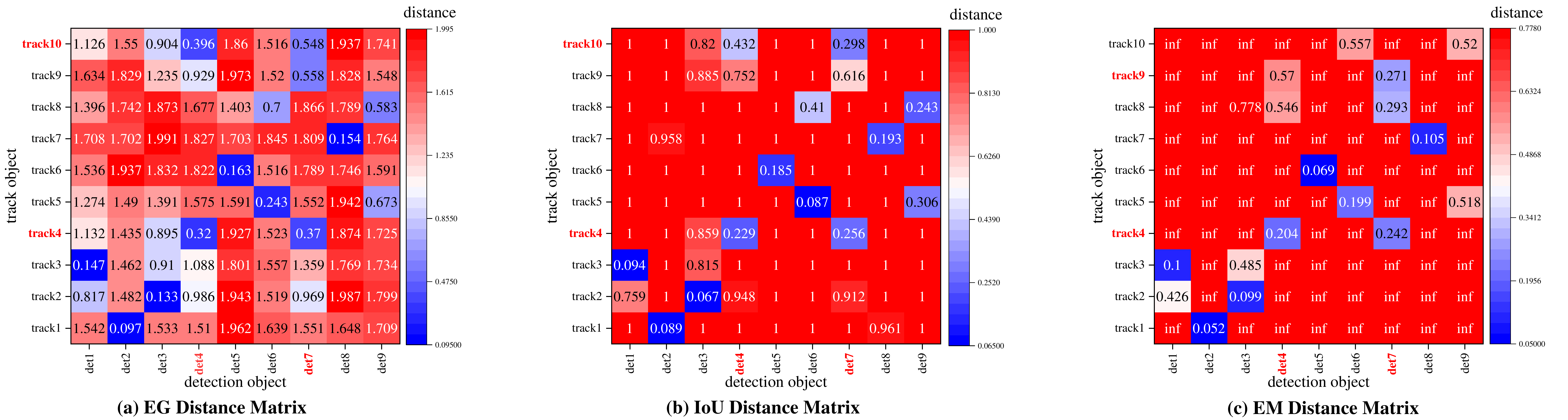}
\caption{The Example of heatmaps for different association matrices in frame 560 of MOT17 sequence 11. (a) shows our EG matrix, which combines the embedding cosine distance and the Giou distance. (b) shows the IoU distance matrix, \emph{i.e.} the detection-based methods. (c) shows the EM matrix, which usually combines the motion distance and the embedding cosine distance, \emph{i.e.} the JDE-based methods. In these heatmaps, the red cells indicates THAT the similarity distance between detection targets and tracking targets is farther, and the blue cells show THAT the similarity distance is closer.}
\label{fig:heatmap}
\end{figure}

\begin{figure}
\centering
\includegraphics[height=7cm]{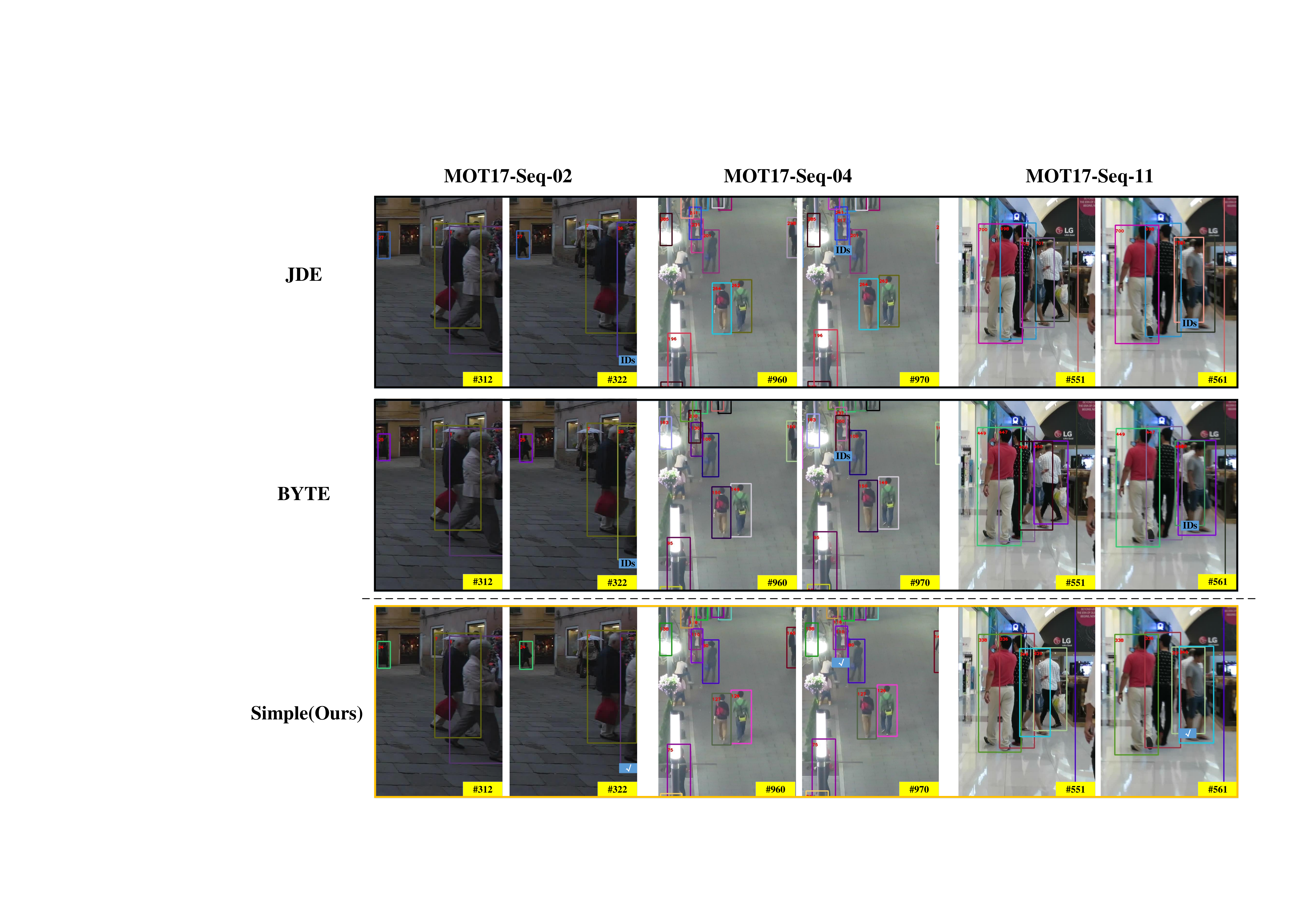}
\caption{Robustness of our tracking strategy compared to BYTE and JDE-based methods. Boxes with the same color indicate that the tracking targets have the same identity, IDs indicates that the tracking targets have switched their identities. The checkmark indicates that the identity of the target has not changed.}
\label{fig:IDsw}
\end{figure}

When objects are occluded due to interlacing, it will produce confusing sets, which are difficult to allocate correctly, \emph{e.g.} the set \{det4, det7, track4, track10\} in Figure \ref{fig:heatmap} (a) and (b), and the set \{det4, det7, track4, track9\} in Figure \ref{fig:heatmap} (c). When assigning these confusing sets, the inaccurate similarity distance leads to tracking failure. Based on the Hungarian method, the IoU distance matrix tends to match det4 with track4 and det7 with track10, and the EM distance matrix tends to match det4 with track4 and det7 with track9. Both of them lead to the target identity switching as shown in MOT17-Seq-11 of Figure \ref{fig:IDsw}. The principal reason for these matching failure is the inaccurate prediction from the Kalman filter as the time of target loss becomes longer. Clearly, this results in inaccurate IoU distance and motion information distance, which leads to the problem of linear allocation errors.

To solve this problem, we propose the EG matrix which utilizes the embedding cosine distance for long-range tracking of targets and the Giou distance for limiting the matching range of embedding. To illustrate the robustness of the EG matrix, we apply it to 5 different JDE-based methods. As can be seen in Section 4.3, our implements obtain improvements in MOT metrics, including tracking speed, HOTA, IDF1, and IDsw metrics.  

To further explore the good property of the EG matrix, we propose a simple tracking framework named SimpleTrack. In this framework, we design a bottom-up branch to represent Re-id features. Different from the fusion method of detection features, it pays more attention to the high-level semantic layers. For the tracking part of SimpleTrack, we propose a novel tracking retrieval mechanism and design a new tracking strategy based on our EG matrix. The experimental results show that our tracking strategy can surpass the JDE-based methods in most metrics, including tracking speed. Compared with the current SOTA method BYTE, our tracking strategy can also improve the performances of HOTA, IDF1, and IDsw metrics. 

%------------------------------------------------------------------------- 
\section{Related Work}
\subsection{Joint Detection and Embedding}
JDE-based methods typically employ a single network to directly predict detection and appearance features\cite{liang2020rethinking,liang2021one,lu2020retinatrack,pang2021quasi,wang2020towards,zhang2021voxeltrack,zhang2021fairmot}. In general, these methods employ the single backbone to predict both object bounding boxes and appearance features. However, the competitive relationship between detection and identification hurts the optimization procedure in the multi-task learning of object detection and appearance feature extraction.

Recently, to tackle this problem, CSTrack\cite{liang2020rethinking} first designs a decoupling module to enhance the learned representation for both object detection and appearance identification. RelationTrack\cite{yu2022relationtrack} uses a channel attention mechanism to decouple detection and Re-identity. Different from CSTrack and RelationTrack, the decoupling strategy adopted in our SimpleTrack focuses on the essence of the appearance feature. We start decoupling from the feature layer fusion of the network. In contrast to the detection feature fusion, we adopt a bottom-up fusion method.
\subsection{Similarity Matrices}
Location, motion and appearance are the most common cues in Multi-object tracking. They are also combined together for the linear assignment. Detection-based methods\cite{zhang2021bytetrack} utilize the IoU distance as the similarity matrix and the tracking accuracy mainly depends on the detector. SORT\cite{bewley2016simple} fuses position and motion cues as the similarity matrix, which can achieve good results in short-range matching. DeepSORT\cite{wojke2017simple} improves the long-range tracking ability of trackers by merging appearance and motion cues, which is usually used in JDE-based methods. 

All these methods use location cue or fuse appearance and motion information as the similarity matrix. However, we design the similarity matrix combined with appearance and location information and use the Giou distance matrix as the location cue instead of the common IoU matrix.
\subsection{Tracking Strategy}
The assignment problem of target tracking and detection can be solved by Hungarian Algorithm based on different similarity matrices. SORT associates the detection objects with the tracking objects by once matching. DeepSORT adopts a cascade matching method that reduces unmatched tracking targets. MOTDT\cite{chen2018real} first uses the appearance similarity matrix and the IoU distance matrix as the similarity matrix for cascade matching respectively.

Recently, BYTETrack\cite{zhang2021bytetrack} proposes to use low-confidence detection results for secondary matching, which reduces the problem of target detection failure due to occlusion. Thereby, the occurrence of long-range tracking could be declined, making the linear assignment based on the IoU distance matrix more effective. MAA\cite{stadler2022modelling} adopts different strategies for the blurred detection targets and tracking targets in the similarity matrix. The method can alleviate the inaccuracy of similarity distance caused by the ambiguous targets. Both of them are aim to make up for the shortcomings of the similarity matrix. Based on the idea of BYTE\cite{zhang2021bytetrack}, we redesign the similarity matrix for the JDE-based method and construct a new matching strategy.

\section{SimpleTrack}
\begin{figure}
\centering
\includegraphics[height=3.5cm]{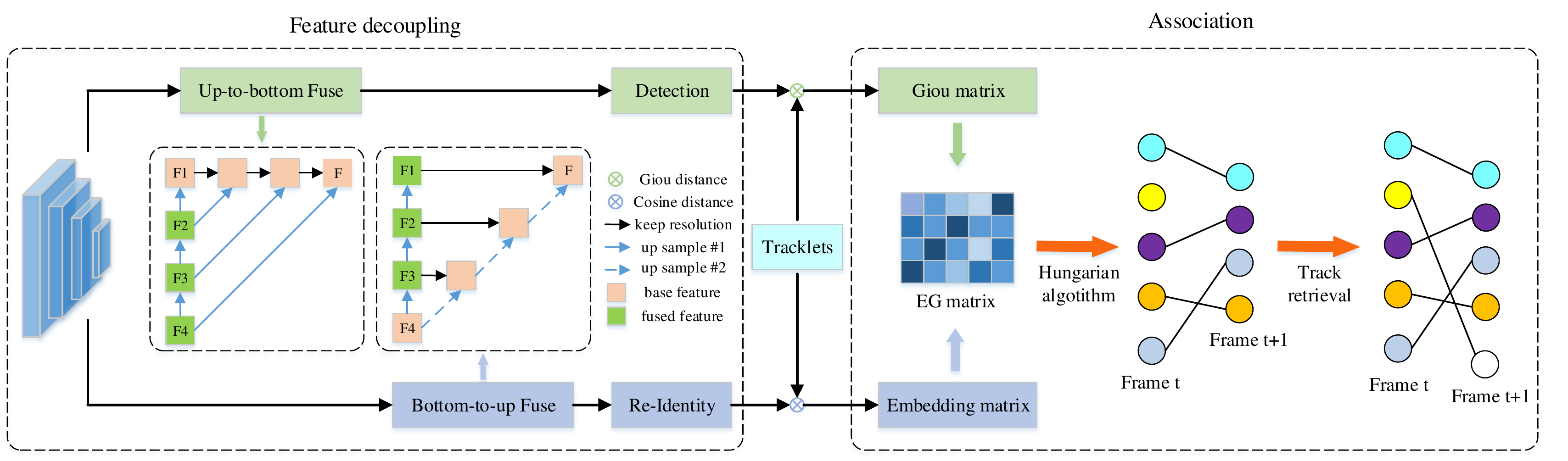}
\caption{The overall pipeline of SimpleTrack. The input image is first fed to a backbone network to extract high-resolution feature maps. Then we use different feature fusion methods for detection and Re-identity separately, and combine the embedding and Giou distance matrix as the similarity matrix. At the end of the association phase, the tracking retrieval mechanism is used to recover the undetected targets.}
\label{fig:framework}
\end{figure}

In this section, we present the technical details of SimpleTrack, as illustrated in Figure \ref{fig:framework}. It is composed of the feature decoupling, the similarity matrix as well as the tracking strategy.  
\subsection{Feature Decoupling}
We adopt DLA-34 as backbone in order to strike a good balance between accuracy and speed. For feature decoupling, we employ different feature fusion methods for detection and ReID representation. As illustrated in Figure \ref{fig:framework}, for the detection branch, the feature fusion method still adopts the structure of IDA-up in Fairmot\cite{zhang2021fairmot}. We call it the up-to-bottom fusion method based on low-level feature maps and continuously fusing higher-level feature maps.

However, Re-ID features tend to learn higher-level semantic features to distinguish different features among homogeneous objects. Therefore, we take a simple bottom-up approach to fusing feature maps. Denote the input feature maps by $\mathbf{F}=\{{\mathbf{F_i}}\}_{i=1}^N$, where $N$ is the number of feature layers of different resolutions extracted by the backbone network. Then, the process of bottom-up fusion method can be expressed as
\begin{align}
 \{\mathbf{\hat{F}}_i\}_{i=N}^1 =
\begin{cases} 
\mathbf{F}_i,  & \text{if }i=N \\
\mathbf{F}_i\cdot\sigma(Conv_{1\times1}(UpSample(\mathbf{\hat{F}}_{i+1}))), & \text{otherwise}
\end{cases}
\end{align}
where $UpSample(\cdot)$ represents an upsampling operation composed of the deformable convolution and the deconvolution, $Conv_{1\times1}$ denotes a $1\times1$ convolution layer for changing channels of features, $\sigma(\cdot)$ represents the Sigmoid activation layer.

It could be observed from Equation (1) that the fusion process is from bottom to top, and the previously fused feature map guides the lower-level feature map until the final fusion result is obtained. As will be shown by the experimental results in Section 4, the computational cost required by this fusion method is minimal.
\subsection{Embedding and Giou Matrix}
The similarity matrix is usually constructed from location, motion and appearance information. Let $\mathbf{L}$, $\mathbf{M}$, $\mathbf{E}$ denote  the location distance matrix, the motion distance matrix and the appearance distance matrix respectively. We fuse $\mathbf{L}$ and $\mathbf{E}$ as the similarity matrix called the EG matrix. And, $\mathbf{L}$ can be represented as
\begin{align}
\mathbf{L} = \mathbf{1} - (\frac{\mid{\mathbf{A}\cap{\mathbf{B}}}\mid}{\mid{\mathbf{A}\cup{\mathbf{B}}}\mid}-\frac{\mid{\mathbf{C}\backslash(\mathbf{A}\cup(\mathbf{B}))}\mid}{\mid{\mathbf{C}}\mid})
\end{align}
where $\mathbf{A}$ and $\mathbf{B}$ represent the bounding boxes of the tracking objects and the bounding boxes of the detection objects respectively, and $\mathbf{C}$ is the minimum enclosing rectangle sets of the above bounding boxes. 

$\mathbf{E}$ can be represented as
\begin{align}
\mathbf{E} = \frac{\mathbf{O}_e^1\cdot{\mathbf{O}_e^2}}{\parallel{\mathbf{O}_e^1}\parallel\parallel{\mathbf{O}_e^2}\parallel}
\end{align}
where $\mathbf{O}_e^1$ and $\mathbf{O}_e^2$ represent different appearance embedding vectors.

Note that the matrix $\mathbf{L}$ in Equation (2) is actually the Giou distance matrix and that the matrix $\mathbf{E}$ in Equation (3) defines the Cosine distance matrix. Then, the Embedding and Giou matrix, which is also denoted as $\mathbf{EG}$, can be represented as 
\begin{align}
\mathbf{EG} = \lambda_1\mathbf{E}+\lambda_2\mathbf{G}
\end{align}
where $\lambda_1=1.0$ and $\lambda_2=0.5$ represent two hyperparameters, $\mathbf{G}$ denotes Giou distance matrix and $\mathbf{G}=\mathbf{L}$.
\subsection{Tracking Strategy in SimpleTrack}
Inspired by BYTE\cite{zhang2021bytetrack}, we develop a tracking strategy based our EG matrix. As shown in Algorithm 1, we follow the idea of secondary matching with low confidence detection adopted in BYTE, and use the EG matrix to replace the similarity matrix in the cascade matching. In addition, after the secondary matching, we utilize the cosine distance to retrieve the unmatched tracklets.

In the retrieving process presented by red texts in Algorithm 1, we use Kalman filter to predict the center point position of the unmatched tracking targets. In order to compensate for the drift of Kalman filter, we select the appearance embedding vectors in the 3$\times$3 range around the prediction center point. Afterward, the minimum cosine distance between these vectors and the embedding vector memorized in the unmatched tracking object can be calculated. If the distance is less than the threshold, the tracklet is retrieved. By tracking retrieval mechanism, we can recover the occluded(failed) detection boxes by using the predictions of Kalman filter.

\begin{algorithm}
\caption{Pseudo-code of SimpleTrack}
\LinesNumbered
\KwIn{A video sequence $V$; object detector $Det$; Kalman Filter $KF$; detection score threshold $\tau_{high}$, $\tau_{low}$; tracking score threshold $\epsilon$; tracking retrieval threshold $\epsilon_r$}
\KwOut{Tracks $\mathcal{T}$ of the video}
% Initialization: $\mathcal{T}\leftarrow\emptyset$\;
\For{frame $f_k$ in $V$}
{
    $\mathcal{D}_{k}\leftarrow{Det(f_k)}$\;
    $\mathcal{D}_{high}\leftarrow\emptyset$\;
    $\mathcal{D}_{low}\leftarrow\emptyset$\;
    \For {$d$ in $\mathcal{D}_k$}
    {
        \If{$d.score>\tau_{high}$}
        {
        $\mathcal{D}_{high}\leftarrow\mathcal{D}_{high}\cup$\{$d$\}\;
        }
        \If{$d.score>\tau_{low}$}
        {
        $\mathcal{D}_{low}\leftarrow\mathcal{D}_{low}\cup$\{$d$\}\;
        }
    }
    \For{$t$ in $\mathcal{T}$}
    {
        $t\leftarrow{KF(t)}$\;
    }
    \tcp{first association with EG matrix}
    \textcolor{blue}{Associate $\mathcal{T}$ and $\mathcal{D}_{high}$ using EG matrix}\;
    $\mathcal{D}_{remain}\leftarrow$remaining object boxes from $\mathcal{D}_{high}$\;
    $\mathcal{T}_{remain}\leftarrow$remaining tracks from $\mathcal{T}$\;
    \tcp{second association with EG matrix}
    \textcolor{blue}{Associate $\mathcal{T}$ and $\mathcal{D}_{low}$ using EG matrix}\;
    $\mathcal{T}_{re-remain}\leftarrow$remaining tracks from $\mathcal{T}$\;
    \textcolor{red}{\tcp{tracking retrieval}
    \For{$t_u$ in $\mathcal{T}_{u}$}
    {
        Find surrounding embedding vectors $\mathbf{E_d}$ with the center point of $t_u$ in the detection frame\;
        Select the most similar appearance embedding vector $\mathbf{E_d^{s}}$ based on the cosine similarity\;
        \If{$\mid{E_u-E_d^{s}}\mid<\epsilon_r$}
        {
          $\mathcal{T}_{reback}\leftarrow{t_u}$\;
        }
    }}
    \tcp{delete unmatched tracks}
    $\mathcal{T}\leftarrow\mathcal{T}\backslash\mathcal{T_{re-remain}}$\;
    \tcp{initialize new tracks}
    \For{$d$ in $\mathcal{D}_{remain}$}
    {
        \If{$d.score>\epsilon$}
        {
          $\mathcal{T}\leftarrow\mathcal{T}\cup{\{d\}}$ \;
        }
    }
}
\textbf{final} \;
\textbf{return} $\mathcal{T}$;
\end{algorithm}

\section{Experiments}
\subsection{Datasets and Metrics}
\subsubsection{Datasets.}
We evaluate SimpleTrack on private detection tracks of MOT17\cite{milan2016mot16} and MOT20\cite{dendorfer2020mot20} datasets. The former contains 14 different video sequences for multi-target tracking recorded by fixed or moving cameras. The latter consists of 8 video sequences with fixed camera focusing on tracking in very crowded scenes, 4 for training and testing each. For ablation studies, we follow \cite{saleh2021probabilistic,shan2020tracklets,wang2021multiple,wu2021track,zhou2020tracking} and split the train set into two parts for ablative experiments as the annotations of the test split are not publicly available. We fuse the CrowdHuman\cite{shao2018crowdhuman} and MOT17 half as the training dataset for ablation experiments following \cite{sun2020transtrack,wu2021track,zeng2021motr,zhang2021bytetrack,zhou2020tracking}. We add the ETH\cite{ess2008mobile}, CityPerson\cite{zhang2017citypersons}, CalTech\cite{dollar2009pedestrian}, CUHK-SYSU\cite{xiao2017joint} and PRW\cite{zheng2017person} datasets for training following \cite{liang2020rethinking,wang2020towards,zhang2021fairmot} when testing on the test set of MOT17.
\subsubsection{Evaluation Metrics.}
To evaluate tracking performance, we use TrackEval\cite{luiten2020trackeval} to evaluate all metrics, including MOTA\cite{bernardin2006multiple}, IDF1\cite{ristani2016performance}, false positives (FP), false negatives (FN), identity switches (IDSW), and recently proposed HOTA\cite{luiten2021hota}. HOTA can comprehensively evaluate the performance of detection and data association. IDF1 focuses more on the association performance and MOTA evaluates the detector ability and focuses more on detection performance. 
\subsection{Implementation Details}
\subsubsection{Tracker.}
In the tracking phase, the default high detection score threshold $\tau_{high}$ is 0.3, the low threshold $\tau_{low}$ is 0.2, the trajectory initialization score $\epsilon$ is 0.6, and the trajectory retrieval score $\epsilon_{r}$ is 0.1, unless otherwise specified. In the linear assignment step, for the high-confidence detection, the assignment threshold is 0.8, and for the low-confidence detection, the assignment threshold is 0.4.
\subsubsection{Detector and Embedding.}
We use SimpleTrack to extract the location features and appearance features of objects. For SimpleTrack, the backbone is DLA-34 which initializes weights with COCO-pretained model. The training schedule is 30 epochs on the combination of MOT17, CrowdHuman, and other datasets mentioned above. The input image size is 1088$\times$608. Rotation, scaling and color jittering are adopted as data augmentation techniques during our training phase. The model is trained on 4 NVIDIA TITAN RTX with a batch size of 32. The optimizer is Adam and the initial learning rate is set to $2\times10^{-4}$ which decays to $2\times10^{-5}$ in the 20 epoch. The total training time is about 25 hours. FPS is measured with a single NVIDIA RTX2080Ti and the batch size is set to 1.
\subsection{Ablation Studies}
\subsubsection{Ablation on SimpleTrack.}
The innovation of SimpleTrack is mainly composed of bottom-up decoupling, EG similarity matrix and tracking retrieval. We conduct ablation experiments on the MOT17 validation set for these three modules. The results are shown in Table \ref{table:simpleablation}. It can be observed that adding bottom-up decoupling to Fairmot increases IDF1 and MOTA. In addition, after replacing the similarity matrix of JDE-based methods with the EG matrix, the strategy improves IDF1 from 76.1 to 78.1, MOTA from 71.4 to 72.5, HOTA from 60.2 to 61.5 and decreases IDs from 451 to 186. After further adding the tracking retrieval mechanism, the IDF1 metric increases from 78.1 to 78.5, HOTA from 61.5 to 61.7 and IDs metric decreases from 186 to 182. These results prove that the modules proposed in SimpleTrack are necessary and effective.
\setlength{\tabcolsep}{4pt}
\begin{table}
\begin{center}
\caption{Ablation experiment on SimpleTrack. $\checkmark$ denotes adding this module to the baseline which is Fairmot. BU-D, EG and TR stand for bottom-up decoupling, EG similarity matrix and tracking retrieval strategy respectively. The best results are show in \textbf{bold}.}
\label{table:simpleablation}
\begin{tabular}{llllllllll}
\hline\noalign{\smallskip}
\multicolumn{3}{l}{Model Settings} &\multicolumn{7}{l}{Evaluation indecators} \\
\hline\noalign{\smallskip}
BU-D & EG & TR  & \makecell[c]{IDF1$\uparrow$} &  \makecell[c]{MOTA$\uparrow$} &   \makecell[c]{HOTA$\uparrow$} &   \makecell[c]{IDs$\downarrow$} &   \makecell[c]{FP$\downarrow$} &   \makecell[c]{FN$\downarrow$} &   \makecell[c]{FPS$\uparrow$}\\
\hline\noalign{\smallskip}
& & & 75.6 & 71.1 & - & 327 & - & - & -\\
\makecell[c]{\checkmark} & & & 76.1 & 71.4 & 60.2 & 451 & 3319 & 11655 & 19.7\\
\makecell[c]{\checkmark}& \makecell[c]{\checkmark} & & 78.1 & 72.5 & 61.5 & 186 & 3260 & \textbf{11430} & \textbf{24}\\
\makecell[c]{\checkmark}& \makecell[c]{\checkmark} &\makecell[c]{\checkmark} & \textbf{78.5} & \textbf{72.5} & \textbf{61.7} & \textbf{182} & \textbf{3212} & 11456 & 23.8\\
\hline
\end{tabular}
\end{center}
\end{table}
\setlength{\tabcolsep}{1.4pt}
\subsubsection{Analysis on the Similarity Matrix.}
We employ different distance matrices as the similarity measure and evaluate their data association ability on the half validation set of MOT17. It can be obtained from Table \ref{table:similaritycompare}, only using the Giou or embedding matrix for data association does not perform well. Besides, the table show that the combination of embedding matrix and IoU matrix can improve the association effect but reduce the result of MOTA. Compared with the IoU matrix used in detection-based methods, our EG matrix improves the IDF1 from 75.7 to 78.5, HOTA from 60.4 to 61.7 and decreases IDs from 285 to 182. Compared with the embedding and motion matrix used in JDE-based methods, our EG matrix improves both the MOT metrics and tracking speed.
\setlength{\tabcolsep}{4pt}
\begin{table}
\begin{center}
\caption{Data association comparison of different similarity matrices. The best results are show in \textbf{bold}.}
\label{table:similaritycompare}
\begin{tabular}{llllllll}
\hline\noalign{\smallskip}
Similarity Matrix & \makecell[c]{IDF1$\uparrow$} &  \makecell[c]{MOTA$\uparrow$} &   \makecell[c]{HOTA$\uparrow$} &   \makecell[c]{IDs$\downarrow$} &   \makecell[c]{FP$\downarrow$} &   \makecell[c]{FN$\downarrow$} &   \makecell[c]{FPS$\uparrow$}\\
\noalign{\smallskip}
\hline
\noalign{\smallskip}
IoU  & 75.7 & 72.5 & 60.4 & 285 & 3510 & 11048 & \textbf{25}\\
GioU  & 66.4 & 70.4 & 54.8 & 378 & 4631 & \textbf{10956} & 23.6\\
Embedding  & 64.1 & 65 & 53.4 & 749 & 6120 & 12012 & 24.2\\
Embedding and Motion   & 76.1 & 71.4 & 60.2 & 451 & 3319 & 11655 & 19.7\\
Embedding and IoU  & 77.2 & 72.3 & 61.4 & 263 & \textbf{2560} & 12144 & 24\\
Embedding and GioU   & \textbf{78.5} & \textbf{72.5} & \textbf{61.7} & \textbf{182} & 3212 & 11456 & 23.8\\
\hline
\end{tabular}
\end{center}
\end{table}
\setlength{\tabcolsep}{1.4pt}
\subsubsection{Applications on other JDE-based Trackers.}
We apply our EG matrix on 5 different JDE-based trackers, including JDE\cite{wang2020towards}, FairMOT\cite{zhang2021fairmot}, CSTrack\cite{liang2020rethinking}, TraDes\cite{wu2021track} and QuasiDense\cite{pang2021quasi}. Among these trackers, JDE, FairMOT, CSTrack, TraDes merge the motion and Re-ID similarity and the first three methods follow the same fusion strategy. QuasiDense uses Re-ID similarity alone. It can be observed from Table \ref{table:othertrack} that using the EG matrix instead of the EM matrix can enhance the tracking performance and improve the tracking speed. Taking the JDE\cite{wang2020towards} method as an example, only using EG matrix to replace the EM matrix can improve the HOTA from 50.1 to 50.9, IDF1 from 63 to 64.4, MOTA from 59.3 to 59.5, FPS from 16.64 to 21.29 and decreases the IDs from 621 to 558. Combined with the BYTE strategy, our EG matrix still improves the HOTA from 50.4 to 50.9, IDF1 from 64.1 to 64.4, FPS from 18.52 to 25.48 and decreases the IDs from 437 to 388.
\setlength{\tabcolsep}{4pt}
\begin{table}
\begin{center}
\caption{Results of applying SimpleTrack to 5 different JDE-based trackers on the MOT17 validation set. Blue represents the tracking method using only the EG matrix, and red represents the tracking method combining the EG matrix and BYTE.}
\label{table:othertrack}
\begin{tabular}{llllllll}
\Xhline{1.0pt}\noalign{\smallskip}
Method  & \makecell[c]{Similarity} & \makecell[c]{w/BYTE} & \makecell[c]{HOTA$\uparrow$} &  \makecell[c]{IDF1$\uparrow$} &  \makecell[c]{MOTA$\uparrow$} &     \makecell[c]{IDs$\downarrow$}  &   \makecell[c]{FPS$\uparrow$}\\
\noalign{\smallskip}
\hline
\noalign{\smallskip}
JDE\cite{wang2020towards}  & \makecell[c]{EM} & \makecell[c]{-} & 50.1 & 63 & 59.3  & 621 & 16.64\\
  & \makecell[c]{EG} & \makecell[c]{-} & \textcolor{blue}{50.9} & \textcolor{blue}{64.4} & \textcolor{blue}{59.5}  & \textcolor{blue}{558} & \textcolor{blue}{21.29}\\
  & \makecell[c]{EM} & \makecell[c]{$\checkmark$} & 50.4 & 64.1 & 60.2 &  437 & 18.52\\
  & \makecell[c]{EG} & \makecell[c]{$\checkmark$} & \textcolor{red}{50.9} & \textcolor{red}{64.8} & \textcolor{red}{60.1} &  \textcolor{red}{388} & \textcolor{red}{25.48}\\
\noalign{\smallskip}
\Xhline{0.5pt}
\noalign{\smallskip}
FairMOT\cite{zhang2021fairmot}  & \makecell[c]{EM} & \makecell[c]{-} & 57 & 72.4 & 69.1  & 372 & 21.01\\
  & \makecell[c]{EG} & \makecell[c]{-} & \textcolor{blue}{57.5} & \textcolor{blue}{73.3} & \textcolor{blue}{69.5}  & \textcolor{blue}{236} & \textcolor{blue}{25.18}\\
  & \makecell[c]{EM} & \makecell[c]{$\checkmark$} & - & 74.2 & 70.4  & 232 & -\\
  & \makecell[c]{EG} & \makecell[c]{$\checkmark$} & \textcolor{red}{58.5} & \textcolor{red}{74.5} & \textcolor{red}{70.6} &  \textcolor{red}{188} & \textcolor{red}{24.7}\\
\noalign{\smallskip}
\Xhline{0.5pt}
\noalign{\smallskip}
CSTrack\cite{liang2020rethinking}  & \makecell[c]{EM} & \makecell[c]{-} & 58.7 & 72.0 & 67.9  & 423 & 20.39\\
  & \makecell[c]{EG} & \makecell[c]{-} & \textcolor{blue}{59.3} & \textcolor{blue}{73.0} & \textcolor{blue}{68.2}  & \textcolor{blue}{322} & \textcolor{blue}{24.3}\\
  & \makecell[c]{EM} & \makecell[c]{$\checkmark$} & 59.8 &73.9 & 69.2 & 298 & 20.72\\
  & \makecell[c]{EG} & \makecell[c]{$\checkmark$} & \textcolor{red}{60.0} & \textcolor{red}{73.8} & \textcolor{red}{69.6} &  \textcolor{red}{249} & \textcolor{red}{24.25}\\
\noalign{\smallskip}
\Xhline{0.5pt}
\noalign{\smallskip}
TraDes\cite{wu2021track}  & \makecell[c]{EM} & \makecell[c]{-} & 58.6 & 71.7 & 68.3  & 293 & 15.8\\
& \makecell[c]{EM} & \makecell[c]{$\checkmark$} & 58.4 & 71.2 & 68.9  & 263 & 16.22\\
& \makecell[c]{EG} & \makecell[c]{$\checkmark$} & \textcolor{red}{59.0} & \textcolor{red}{71.5} & \textcolor{red}{68.5}  & \textcolor{red}{483} & \textcolor{red}{16.5}\\
\noalign{\smallskip}
\Xhline{0.5pt}
\noalign{\smallskip}
QuasiDense\cite{pang2021quasi}  & \makecell[c]{EM} & \makecell[c]{-} & 56.2 & 67.7 & 67.1  & 386 & 4.1\\
  & \makecell[c]{EM} & \makecell[c]{$\checkmark$} & 58.5 & 71.9 & 67.4  & 295 & 4.8\\
  & \makecell[c]{EG} & \makecell[c]{$\checkmark$}& \textcolor{red}{57.9} & \textcolor{red}{70.9} & \textcolor{red}{67.5}  & \textcolor{red}{252} & \textcolor{red}{4.8}\\
\Xhline{1.0pt}
\end{tabular}
\end{center}
\end{table}
\setlength{\tabcolsep}{1.4pt}
\subsubsection{The Accuracy Compared with other Association Methods.}
We compare SimpleTrack with other association methods, including the recent SOTA algorithm BYTE and the tracking algorithm used in JDE-based methods\cite{zhang2021fairmot,wang2020towards,liang2020rethinking,liang2021one}. As shown in Table \ref{table:trackcompare}, SimpleTrack improves the IDF1 metric of JDE from 76.1 to 78.5, MOTA from 71.4 to 72.5, HOTA from 60.2 to 61.7 and decreases IDs from 451 to 182. Compared with BYTE, we can see that SimpleTrack improves the IDF1 from 75.7 to 78.5, HOTA from 60.4 to 61.7, and decreases IDs from 285 to 182. These demonstrate that our tracking method is more effective than the JDE strategy, and it can improve the accuracy of data association compared to the BYTE strategy.
\setlength{\tabcolsep}{4pt}
\begin{table}
\begin{center}
\caption{Comparison of different association methods on the MOT17 validation set. JDE expresses the tracking strategy employed by \cite{liang2020rethinking,liang2021one,wang2020towards,zhang2021fairmot} and BYTE expresses the tracking strategy employed by \cite{zhang2021bytetrack}. The best results are shown in \textbf{bold}.}
\label{table:trackcompare}
\begin{tabular}{llllllll}
\hline\noalign{\smallskip}
Track Method & \makecell[c]{IDF1$\uparrow$} &  \makecell[c]{MOTA$\uparrow$} &   \makecell[c]{HOTA$\uparrow$} &   \makecell[c]{IDs$\downarrow$} &   \makecell[c]{FP$\downarrow$} &   \makecell[c]{FN$\downarrow$} &   \makecell[c]{FPS$\uparrow$}\\
\noalign{\smallskip}
\hline
\noalign{\smallskip}
JDE   & 76.1 & 71.4 & 60.2 & 451 & 3319 & 11655 & 19.7\\
BYTE  & 75.7 & 72.5 & 60.4 & 285 & 3510 & \textbf{11048} & \textbf{25}\\
SimpleTrack(Ours)  & \textbf{78.5} & \textbf{72.5} & \textbf{61.7} & \textbf{182} & \textbf{3212} & 11456 & 23.8\\
\hline
\end{tabular}
\end{center}
\end{table}
\setlength{\tabcolsep}{1.4pt}
\subsubsection{The Speed Compared with other Association Methods.}
\begin{figure}
\centering
\includegraphics[height=5.2cm]{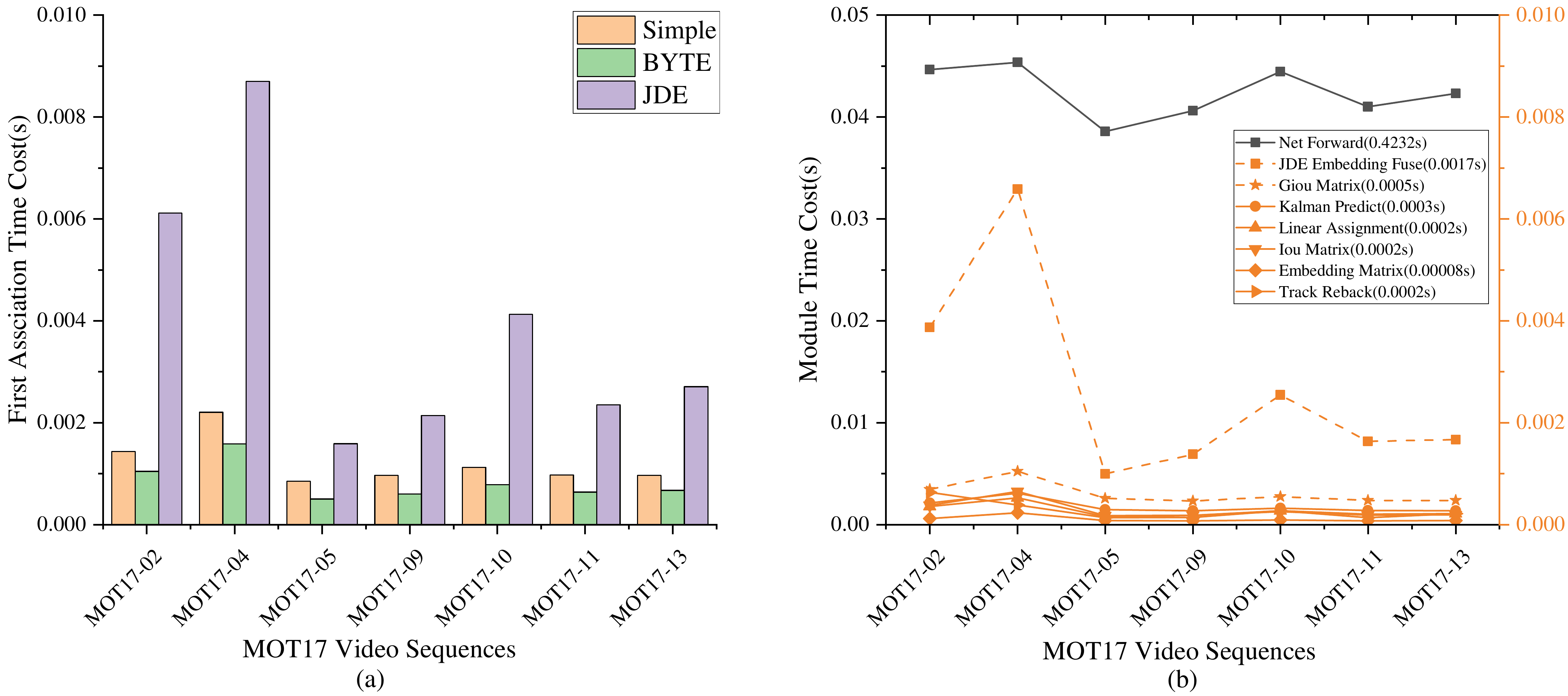}
\caption{Comparison of different tracking algorithm speeds. (a) shows the tracking speed of different tracking algorithms. (b) shows the time-consuming of several main modules in the tracking phase.}
\label{fig:speed}
\end{figure}

From Table \ref{table:trackcompare} and Table \ref{table:similaritycompare}, we can observe that our SimpleTrack algorithm utilizes the embedding information but is still nearly 20\% faster than JDE's tracking strategy. A more detailed comparison of different video sequences can be observed in Figure \ref{fig:speed} (a). It can be observed that our tracking algorithm is only slightly slower than BYTE which does not utilize the embedding information. According to Figure \ref{fig:speed} (b), we can see time-consuming of the main modules in the tracking phase. It shows that the JDE-based tracking strategy spends a lot of time in fusing the embedding and motion information, which is represented by the orange dotted square in Figure \ref{fig:speed} (b). For the EG matrix, we only need to calculate the Giou distance and add it with the embedding distance. The time consumption is represented by the orange dotted star in Figure \ref{fig:speed} (b).
\subsubsection{Comparison with Preceding SOTAs.}
In this part, we compare the performance of SimpleTrack with preceding SOTA methods on MOT17 and MOT20. The results are reported in Table \ref{table:MOT17} and Table \ref{table:MOT20}, respectively. As shown in these two tables, SimpleTrack has come out among the top in various metrics and surpassed the contrasted counterparts by large margins, especially on the HOTA, IDF1 and IDS metrics. Besides, compared with other MOT tracking methods, SimpleTrack has obvious speed advantage.

\subsubsection{Visualization Results.}
\begin{figure}
\centering
\includegraphics[height=16cm]{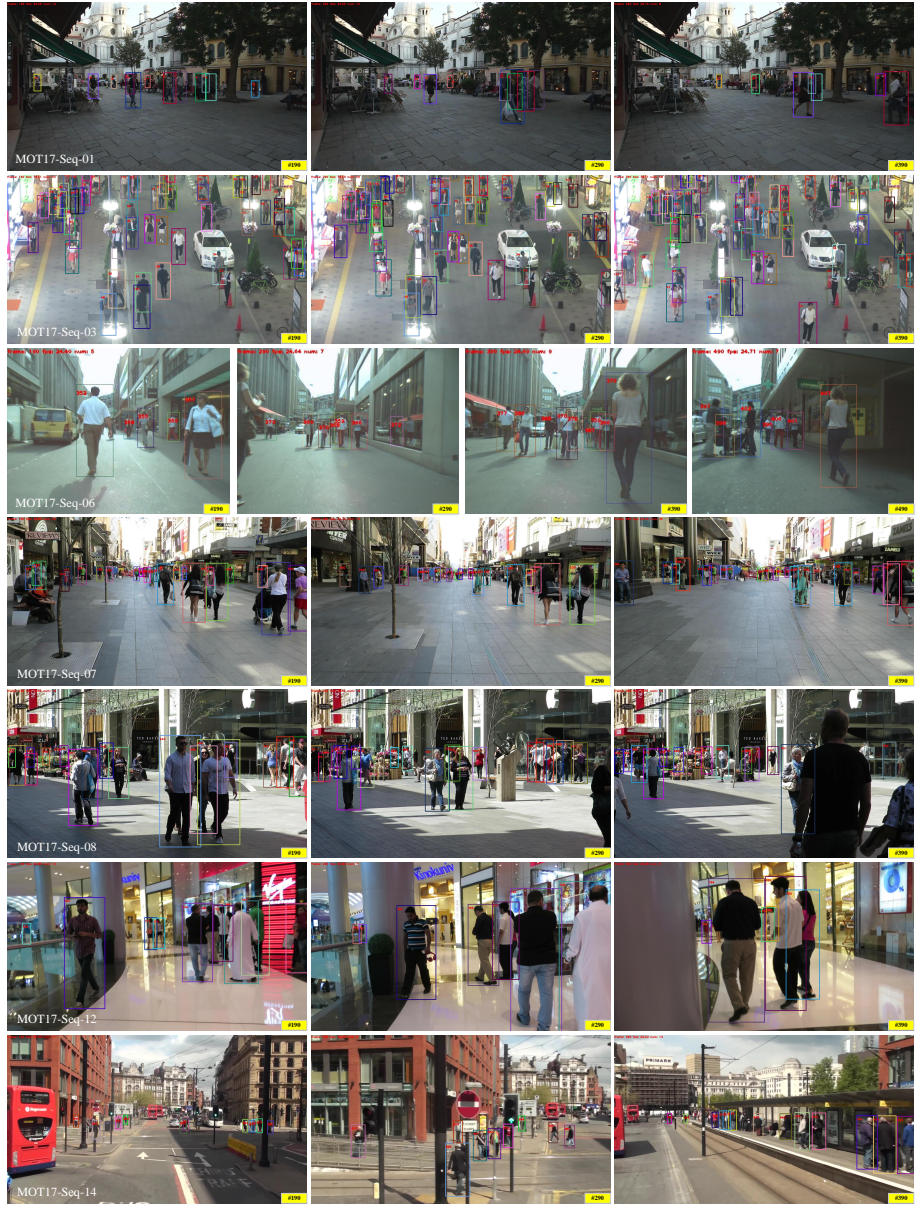}
\caption{Tracking results of SimpleTrack on the MOT17 test dataset.}
\label{fig:MOT17}
\end{figure}

We show some scenarios that are prone to identity switching in Figure \ref{fig:IDsw} which contains 3 sequences from the half validation set of MOT17. We use different tracking strategies to generate the visualization results. It can be observed that SimpleTrack can effectively deal with the identity switching problem caused by the occlusion of the tracking targets. In addition, some tracking examples on the MOT17 test datasets are shown by Figure \ref{fig:MOT17}.
\setlength{\tabcolsep}{4pt}
\begin{table}
\begin{center}
\caption{Comparison of the state-of-the-art methods under the "private detector" protocol on the MOT17 test set. The best results are shown in \textbf{bold}. MOT17 contains rich scenes and half of the sequences are captured with camera motion.}
\label{table:MOT17}
\begin{tabular}{llllllll}
\hline\noalign{\smallskip}
Method & \makecell[c]{HOTA$\uparrow$} & \makecell[c]{IDF1$\uparrow$} &  \makecell[c]{MOTA$\uparrow$} &      \makecell[c]{IDs$\downarrow$} &   \makecell[c]{FP$\downarrow$} &   \makecell[c]{FN$\downarrow$} &   \makecell[c]{FPS$\uparrow$}\\
\noalign{\smallskip}
\hline
\noalign{\smallskip}
TraDes\cite{wu2021track}$\dagger$  & 52.7 & 63.9 & 69.1  & 3555 & 20892 & 150060 & 17.5\\
MAT  & 53.8 & 63.1 & 69.5  & 2844 & 30660 & 138741 & 9.0\\
QuasiDense\cite{pang2021quasi}$\dagger$  & 53.9 & 66.3 & 68.7  & 3378 & 26589 & 146643 & 20.3\\
SOTMOT\cite{han2022mat}  & - & 71.9 & 71.0  & 5184 & 39537 & 118983 & 16.0\\
TransCenter\cite{xu2021transcenter}  & 54.5 & 62.2 & 73.2  & 4614 & 23112 & 123738 & 1.0\\
PermaTrackPr\cite{tokmakov2021learning}  & 55.5 & 68.9 & 73.8  & 3699 & 28998 & 115104 & 11.9\\
TransTrack\cite{sun2020transtrack} & 54.1 & 63.5 & 75.2 & 3603 & 50157 & 86442 & 10.0\\
FUFET\cite{shan2020tracklets} & 57.9 & 68.0 & 76.2 & 3237 & 32796 & 98475 & 6.8\\
FairMOT\cite{zhang2021fairmot}$\dagger$ & 59.3 & 72.3 & 73.7  & 3303 & 27507 & 117477 &18.9\\
CSTrack\cite{liang2020rethinking}$\dagger$ & 59.3 & 72.6 & 74.9 & 3567 & 23847 & 114303 & 15.8\\
Semi-TCL\cite{wang2021joint} & 59.8 & 73.2 & 73.3  & 2790 & 22944 & 124980 & -\\
ReMOT\cite{yang2021remot} & 59.7 & 72.0 & \textbf{77.0} & 2853 & 33204 & \textbf{93612} & 1.8\\
CrowdTrack\cite{stadler2021performance} & 60.3 & 73.6 & 75.6 & 2544 & 25950 & 109101 & -\\
CorrTracker\cite{wang2021multiple}$\dagger$ & 60.7 & 73.6 & 76.5 & 3369 & 29808 & 99510 & 15.6\\
RelationTrack\cite{yu2022relationtrack}$\dagger$ & 61.0 & 74.7 & 73.8  & 1374 & 27999 & 118623 & 8.5\\
SimpleTrack(Ours)$\dagger$ & 61.0 & 75.7 & 74.1  & 1500  & \textbf{17379} & 127053 & \textbf{22.53}\\
SimpleTrack(Ours)* & \textbf{61.6} & \textbf{76.3} & 75.3  & \textbf{1260}  & 22317 & 116010 & -\\
\hline
\multicolumn{4}{l}{* indicates adding linear interpolation} \\
\multicolumn{4}{l}{$\dagger$ indicates JDE-based methods}

\end{tabular}
\end{center}
\end{table}
\setlength{\tabcolsep}{1.4pt}

\setlength{\tabcolsep}{4pt}
\begin{table}
\begin{center}
\caption{Comparison of the state-of-the-art methods under the "private detector" protocol on the MOT20 test set. The best results are shown in \textbf{bold}. The scenes in MOT20 are much more crowded than those in MOT17.}
\label{table:MOT20}
\begin{tabular}{llllllll}
\hline\noalign{\smallskip}
Method &  \makecell[c]{HOTA$\uparrow$} & \makecell[c]{IDF1$\uparrow$} &  \makecell[c]{MOTA$\uparrow$} &     \makecell[c]{IDs$\downarrow$} &   \makecell[c]{FP$\downarrow$} &   \makecell[c]{FN$\downarrow$} &   \makecell[c]{FPS$\uparrow$}\\
\noalign{\smallskip}
\hline
\noalign{\smallskip}
MLT\cite{zhang2020multiplex}  & 43.2 & 54.6 & 48.9  & 2187 & 45660 & 216803 & 3.7\\
FairMOT\cite{zhang2021fairmot}$\dagger$ & 54.6 & 67.3 & 61.8  & 5243 & 103440 & \textbf{88901} & \textbf{13.2}\\
TransCenter\cite{xu2021transcenter} & - & 50.4 & 61.9  & 4653 & 45895 & 146347 & 1.0\\
TransTrack\cite{sun2020transtrack}  & 48.5 & 59.4 & 65.0  & 3608 & 27197 & 150197 & 7.2\\
Semi-TCL\cite{wang2021joint}  & 55.3 & 70.1 & 65.2  & 4139 & 61209 & 114709 & -\\
CorrTracker\cite{wang2021multiple}$\dagger$ & -  & 69.1 & 65.2 & 5183 & 79429 & 95855 & 8.5\\
CSTrack\cite{liang2020rethinking}$\dagger$  & 54.0 & 68.6 & 66.6  & 3196 & 25404 & 144358 & 4.5\\
GSDT\cite{wang2021joint}$\dagger$  & 53.6 & 67.5 & 67.1  & 3131 & 31913 & 135409 & 0.9\\
SiamMOT\cite{liang2021one}$\dagger$  & - & 67.8 & 70.7  & - & 22689 & 125039 & 6.7\\
RelationTrack\cite{yu2022relationtrack}$\dagger$  & 56.5 & 70.5 & 67.2  & 4243 & 61134 & 104597 & 2.7\\
SOTMOT\cite{han2022mat}  & - & \textbf{71.4} & 68.6  & 4209 & 57064 & 101154 & 8.5\\
SimpleTrack(Ours)$\dagger$  & 56.6 & 69.6 & 70.6  & 2,434 & \textbf{18400} & 131209 & 7.0\\
SimpleTrack(Ours)*  & \textbf{57.6} & 70.2 & \textbf{72.6}  & \textbf{1785} & 25515 & 114463 & -\\
\hline
\multicolumn{4}{l}{* indicates adding linear interpolation} \\
\multicolumn{4}{l}{$\dagger$ indicates JDE-based methods}
\end{tabular}
\end{center}
\end{table}
\setlength{\tabcolsep}{1.4pt}

\section{Conclusions}
We proposed a simple tracking framework called SimpleTrack for data assocaition and multi-object tracking. SimpleTrack was implemented by developing a simple and effective similarity matrix (called EG matrix), which combines the embedding and Giou distance. The proposed EG matrix can improve not only the tracking effect but also the speed of JDE-based tracking methods. Moreover, we also proposed a bottom-up feature fusion module for decoupling Reid and detection tasks, and design a tracking strategy for JDE architecture by combining the BYTE strategy and EG matrix. The results show that SimpleTrack is very competitive, and we hope that the EG matrix will facilitate the development of the JDE-based methods.

\clearpage
%===========================================================

\bibliographystyle{splncs}
\bibliography{SimpleTrack}
\end{document}